\documentclass[lettersize,journal]{Template/IEEEtran}
\usepackage{amsmath,amsfonts}
\usepackage{algorithmic}
\usepackage{array}
\usepackage[caption=false,font=normalsize,labelfont=sf,textfont=sf]{subfig}
\usepackage{textcomp}
\usepackage{stfloats}
\usepackage{url}
\usepackage{verbatim}
\usepackage{graphicx}
\def\BibTeX{{\rm B\kern-.05em{\sc i\kern-.025em b}\kern-.08em
    T\kern-.1667em\lower.7ex\hbox{E}\kern-.125emX}}
\usepackage{balance}

\newcommand\Colored[2]{#2} 
\newcommand\ColorA[1]{\Colored{0.0,0.2,0.9}{#1}} 
\newcommand\ColorB[1]{\Colored{0.8,0,0}{#1}} 

\usepackage{subfiles}
\usepackage{algorithm}
\usepackage{algorithmic}
\usepackage{amsmath}
\usepackage{amsfonts} 
\usepackage{amsthm}
\usepackage{cases}
\usepackage[colorlinks=true,citecolor=cyan]{hyperref}
\usepackage{color}
\usepackage{color}
\usepackage{tcolorbox}
\usepackage{empheq}
\usepackage{framed}
\usepackage{lmodern} 

\theoremstyle{definition}
\newtheorem{problem}{Problem}

\newtheorem{remark}{Remark}
\newtheorem{theorem}{Theorem}
\newtheorem{example}{Example}

\newcommand\Strong[1]{\textbf{\underline{#1}}}

\newcommand\RR{\mathbb{R}}
\newcommand\PP{\mathbb{P}}

\newcommand\Range[2]{\{#1,\ldots,#2\}}

\newcommand\DKL{D_\mathrm{KL}}

\newcommand\NominalInput{u_\mathrm{nom}}
\newcommand\StateSet{\mathcal{X}}
\newcommand\SafeSet{\mathcal{S}}
\newcommand\DangerSet{\bar{\mathcal{S}}}

\newcommand\TokenSet{\mathcal{T}} 
\newcommand\TextSet{\StateSet}

\newcommand\AllowedSet{\mathcal{A}}
\newcommand\DisallowedSet{\mathcal{D}}

\begin{document}
\title{Control Barrier Function for Aligning Large Language Models}
\author{Yuya Miyaoka, Masaki Inoue}

\author{Yuya Miyaoka, Masaki Inoue, \IEEEmembership{Member, IEEE}
\thanks{This work was supported by the Grant-in-Aid for Scientific Research (B), No.~20H02173 and 25K01254 from JSPS.}
\thanks{Y. Miyaoka and M. Inoue are with the Department of Applied Physics and Physico-Informatics, Keio University, 
Yokohama 223-8522, Japan (e-mail:miyaoka.yuya@keio.jp, minoue@appi.keio.ac.jp).}
}


\maketitle

\begin{abstract}
This paper proposes a control-based framework for aligning large language models (LLMs) by leveraging a control barrier function (CBF) to ensure user-desirable text generation. 
The presented framework applies the CBF safety filter to the predicted token generated from the baseline LLM, to intervene in the generated text.
The safety filter includes two significant advantages:
this safety filter is an add-on type, allowing it to be used for alignment purposes without fine-tuning the baseline LLM,
and if there is an evaluation model regarding the desired alignment, it can be directly applied to the filter design.
The overall text-generation system is implemented with open-source language models, aiming to generate positive text.
\end{abstract}

\begin{IEEEkeywords}
Control Barrier Function, Large Language Models, Alignment, Safe Control
\end{IEEEkeywords}


\section{Introduction}\label{Main:I}
While large language models (LLMs) are known to have strong language understanding, reasoning and writing abilities, they can also generate harmful, biased, toxic, or unethical content \cite{Shen23_LLMAlignment, Minaee24_LLM}.
Alignment of LLMs ensures that they generate content that is ``desirable'' for user, meaning that the content is ethical and safe.
Various approaches for LLM alignment have been presented (see the literature \cite{Shen23_LLMAlignment, Minaee24_LLM, Wang23_LLMAlignment} and reference therein).

The major approach to LLM alignment is reinforcement learning from human feedback (RLHF, \cite{Ouyang22}),
where a reward model is constructed by human feedback and then used for the training of LLMs.
Variants of RLHF methods are also proposed, such as Safe-RLHF by \cite{Dai24}, SENSEI by \cite{Ruibo22}, and f-DPG by \cite{Go23},
and their implementations are presented, such as training pre-trained LLMs \cite{Bai22, Chunting23}.
Collecting human feedback with data is time-consuming and expensive.
To overcome this drawback, reinforcement learning from AI feedback (RLAIF) instead of using human feedback is presented by \cite{Bai22b}.
Furthermore, to reduce the computational cost and enhance the stability of training, direct preference optimization (DPO) is proposed by \cite{Rafael23}, where the alignment data is directly used for training LLMs without accessing the reward model.
A common feature of alignment methods like RLHF, variants of RLHF, and DPO is that they update LLMs' model parameters.
An alternative approach for LLM alignment is to directly intervene in the input prompts, rather than updating the model parameters.
In-context learning (ICL, \cite{Dong24_ICL}) is a major approach for intervening in the input prompt.
In ICL, a few pairs of input prompts and output are provided as demonstrations to instruct the LLMs on the task \cite{Tom20, Zhao24}.

\begin{figure}[t]
    \centering
    \includegraphics[width=1\linewidth]{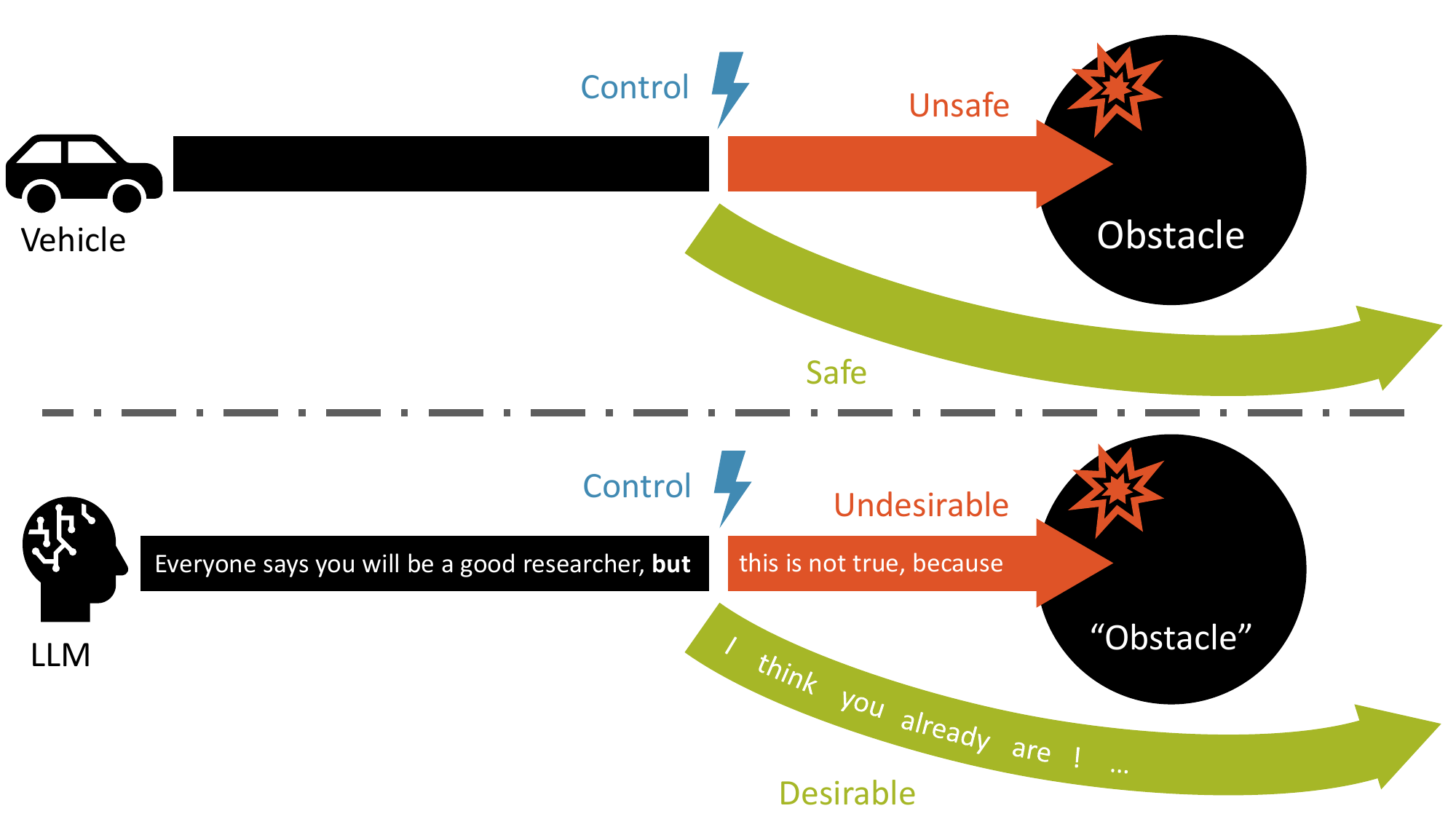}
    \caption{Safe Control of LLM.
    Top: Collision avoidance in a vehicle control system,
    Bottom: \textit{Collision avoidance} in text-generation by LLMs.}
    \label{F:I.Concept}
\end{figure}

\ColorA{The intervention-based approach to LLM alignment, which focuses on preventing undesirable outputs, is analogous to the problem of collision avoidance. The collision avoidance task is the most fundamental control problem in systems engineering.}%
Just as the vehicle's trajectory is intervened to avoid collisions, LLM's output can be intervened to prevent undesirable content.
This paper draws an analogy between a vehicle and LLM, as illustrated in Fig.~\ref{F:I.Concept}.
Consider the LLM as an analogy to a vehicle, and the generated text as an analogy to the vehicle's trajectory.
Both vehicle collision avoidance and LLM alignment aim to guide the complex system away from undesirable states by designing proper control strategies.
Vehicle collision avoidance aims to prevent collisions with obstacles by intervening in the vehicle's trajectory.
Similarly, LLM alignment aims to prevent undesirable outputs, such as harmful and unethical content.
To this end, we intervene in the text-generation system to generate the desirable trajectory of the token sequence.

In the control community, various studies are conducted for safety assurance of control systems, including collision avoidance \cite{Dawson23, Adam21_Safety}.
A promising approach for collision avoidance and safe control is the control barrier function (CBF), as studied in \cite{Ames19, Gurriet18, J20, Issei22_ZCBF}.
\ColorA{The theory for CBF is extended to robust CBF \cite{Brett21}, multiple CBF \cite{Axton24}, reinforcement learning integration \cite{Cheng19_CBF}, CBF for time invarying system \cite{Igarashi19_CBF}, and CBF for stochastic systems \cite{Nishimura24, Allan24_CBF}.
}%
In addition to various theoretical advancements, CBF has been applied in diverse areas, such as robotics \cite{Siavash22_CBF_TCST, Cortez21_CBF_TCST, Funada24_CBF_Quadcopter,Cohen24_CBF, Federica22_CBF}, vehicle control \cite{Anil23_CBF_TCST}, battery management \cite{Shuang24_CBF_TCST}, and learning-based system \cite{Cheng19_CBF}.
For example, the works on robotics \cite{Siavash22_CBF_TCST, Cortez21_CBF_TCST} address the human-machine shared control, ensuring the safety of the human space.
This paper is the first to apply the powerful theory of CBF to LLM alignment.

This paper proposes a framework for control-based LLM alignment by applying a safety filter that intervenes in the LLM output to generate user-desirable outcomes.
To this end, we leverage CBF to improve the safety and controllability of the output of LLMs.
We aim to design a CBF-based safety filter that intervenes in the output of LLMs into the user's desired content.
The CBF filter and the baseline LLM constitute a controlled text-generation system, which we call ``CBF-LLM''.


The contributions of this paper are as follows:
\begin{enumerate}
    \item While theoretical analysis of LLMs has been studied in the works by \cite{Aman24} and \cite{Soatto23}, their design methodology has not. This paper develops a design approach for LLMs through the lens of control theory. 
    \item CBF-based safe control has been successfully applied in various domains. This paper introduces a novel application of CBF-based safe control to the domain of LLMs, aiming to enhance the \textit{safety} and reliability of text-generation systems.
    \item The proposed system, CBF-LLM, is realized in an add-on manner to a baseline LLM: an external filter is simply added to the LLMs without accessing and updating their baseline LLM parameters. In this sense, CBF-LLM is broadly applicable to various alignment goals and various LLMs.
    \item In this paper, CBF-LLM is implemented with Llama~3 \cite{Dubey24}, an open-source LLM developed by Meta, and a RoBERTa model in Section~\ref{Main:E}.
\end{enumerate}

The rest of this paper is as follows.
In Section~\ref{Main:P}, the basic theory of CBF and the text-generation system using LLM are reviewed.
In Section~\ref{Main:M}, the concept and design of CBF-LLM are proposed.
In Section~\ref{Main:E}, the implementation of CBF-LLM is presented and the text-generation experiment is conducted.
Finally, in Section~\ref{Main:C}, the conclusion of this paper is presented.

Notation: symbol $V[i]$ represents the $i$-th element of vector $V$.

\section{Preliminaries}\label{Main:P}

\subsection{Control Barrier Function for Safe Control}\label{Main:P.CBF}
Control barrier function (CBF), developed in the control community, provides safety assurance in control systems \cite{Ames19, Gurriet18}.
This subsection briefly reviews CBF and CBF-based safe control.

Consider the following dynamical system to be controlled:
\begin{align}
    \dot x = g(x,u) , \label{E:P.NominalDynamics}
\end{align}
where $x\in\RR^n$ is the state variable of the control target, and $u\in\RR^m$ is the control input applied to the target,
and $g$ is a nonlinear function that represents the system dynamics.

We aim to design the assisted control system with safety assurance.
Specifically, we focus on achieving safe control by filtering nominal control inputs and modifying them into safe control inputs.
We let the safe and unsafe sets be denoted by $\SafeSet\subseteq\RR^n$, and \ColorA{$\DangerSet = \RR^n\setminus \SafeSet \subseteq\RR^n$}, respectively.
Then, the safety means to constrain the state $x$ within the safe set $\SafeSet$, i.e., $x\in\SafeSet$.
We address the problem of designing the following filter $F:\RR^m\to\RR^m$:
\begin{problem}[Safety Filter]
    \ColorA{Assume that the system \eqref{E:P.NominalDynamics} starts in a safe state $x(\tau_0) \in \SafeSet$ at time $\tau_0$.
    }%
    Given a nominal control input $\NominalInput\in\RR^m$,
    find the safety filter \ColorA{$F:\RR^m\times\RR^n\to\RR^m$} such that the system \eqref{E:P.NominalDynamics} with \ColorA{$u=F(\NominalInput,x)$} generates $x( \tau )\in\SafeSet$ for all time \ColorA{$\tau\ge\tau_0$}.
\end{problem}
As a preliminary, we design a continuous function $h:\RR^n\to\RR$, called a ``constraint function'', such that $h(x) \ge 0$ if $x\in\SafeSet$, and $h(x)<0$ if $x\in\DangerSet$ holds.
The safety is equivalent to the constraint: $h(x) \ge 0$.
The safety filter $F$ needs to modify $\NominalInput$ to output $u$ such that $h(x)\ge0$ holds thereafter.

To construct the safety filter $F$ that keeps the safety constraint $h(x)\ge0$,
the control barrier function filter (CBF filter, \cite{Ames19, Gurriet18}) is presented. 
The CBF filter intervenes in the nominal control input $\NominalInput$ to introduce a safe state of the object by finding $u$ as follows:
\ColorA{
\begin{subequations}
\begin{empheq}[left={
    F: \empheqlbrace
}]{alignat=1}
    \min_u &~ (\NominalInput-u)^2 , \label{E:P.CBFObjective}\\
    \text{s.t.} &~ \dot h(x,u) \ge - \alpha_\mathrm{c} (h(x))  ,\label{E:P.CBFConstraint}
\end{empheq}
\end{subequations}
}%
\ColorA{where $\alpha_c:\RR\to\RR$ is a class-$\mathcal{K}$ function which holds $\alpha_\mathrm{c}(0)=0$ and is monotonically increasing.} 
The function $h$ is called a \textit{control barrier function} if there exists $u$ such that the constraint \eqref{E:P.CBFConstraint} holds.
To constrain the control input by the CBF filter, the following theorem on safety assurance holds:
\begin{theorem}[\cite{Ames19}]
    Suppose that \ColorA{$x(\tau_0)\in\SafeSet$, and the control input $u = u(\tau)$ satisfies the CBF constraint \eqref{E:P.CBFConstraint} for all $\tau\ge\tau_0$.
    Then, the state $x(\tau)\in\SafeSet$ holds for all $\tau \ge \tau_0$.}
\end{theorem}
\ColorA{The objective function \eqref{E:P.CBFObjective} ensures that the filtered control input $u$ remains close to the nominal control input $\NominalInput$.}
In this sense, the CBF filter achieves safety by the ``minimum'' intervention.

The CBF filter is capable of applying in discrete-time systems by re-formulating the CBF constraint \eqref{E:P.CBFConstraint} as follows \cite{Zeng21}:
\ColorA{
\begin{align}
\begin{split}
    & \Delta h(x(k),x(k+1)) \\
    = &  h(x(k+1)) - h(x(k)) \ge - \alpha_\mathrm{d} (h(x(k))) ,
\end{split}
    \label{E:P.DiscreteTimeCBFConstraintBase}
\end{align}
}%
where $k$ is a discrete time, \ColorA{and $\alpha_\mathrm{d} : \RR\to\RR$ is a class-$\mathcal{K}$ function that satisfies $0 \le \alpha_\mathrm{d} (h(x)) \le h(x)$ for all $x$}.

\ColorA{
In this paper, we let the function $\alpha_\mathrm{d} (h(k))$ be a linear function, denoting $\alpha_\mathrm{d} (h(k)):= \alpha h(k)$, where $\alpha\in[0,1]$ is a hyperparameter.
Further letting $\gamma:=1-\alpha$, the CBF constraint \eqref{E:P.DiscreteTimeCBFConstraintBase} is rewritten as:
\begin{align}
    h(x(k+1)) \ge \gamma h(x(k)).
\label{E:P.DiscreteTimeCBFConstraint}
\end{align}
}%

\subsection{Large Language Models for Text Generation}\label{Main:P.LLM}
\ColorA{This subsection briefly reviews and analyzes the text generation by large language models (LLMs).
}%
In this paper, ``text'' means the sequence of tokens, and $\TextSet$ denotes the set of all texts.
For example, we let $x=$``It is a nice day.'', meaning $x$ is composed of six tokens: ``It'', ``is'', ``a'', ``nice'', ``day'', and ``.''.
Each token $t$ is identified by a positive integer, i.e., $t\in\Range{1}{N}:=\TokenSet$, here $N$ is the number of tokens the LLM has.
The operator $\oplus$ concatenates text and token to output a text.
For example, ``It is a''$\oplus$``nice''$=$``It is a nice''.

\ColorA{Text generation by LLMs is performed by iteratively adding a new token while considering all given/generated text.}
Let $x_0\in\TextSet$ be the initial text, and $k\in\{0,1,\ldots\}$ be the discrete time, which counts the number of tokens added during the generation. 
Then, the text generation from the initial text $x(0)=x_0$ is expressed by a following discrete-time dynamical system:
\begin{subequations}
\begin{empheq}[left={
    \empheqlbrace
}]{alignat=1}
    P(k) &= G(x(k)), \label{E:P.TokenPrediction} \\
    t^*(k) &= C(P(k)), \label{E:P.TokenSelection} \\
    x(k+1) &= x(k) \oplus t^*(k). \label{E:P.TextConcatenation}
\end{empheq}
\label{E:P.TextGenerationSystem}
\end{subequations}
In the system description, the symbol $G:\TextSet\to(0,1)^N$ represents the token predictor, which is driven based on the input text $x$ to output the token distribution vector $P\in(0,1)^N$ .
Each element of $P$, denoted by $P[t],t\in\TokenSet$ displays how probable that the token $t\in\TokenSet$ can \ColorA{be followed by} the text $x$.
The symbol $C$ is the token selector, which selects the next token $t^*$ that follows $x$ based on the token distribution vector $P$.
In this context, the text $x$ acts as the state value of the text-generation system \eqref{E:P.TextGenerationSystem}. \ColorA{As a new token $t^*$ is added, the time $k$ proceeds and the state $x$ is updated, i.e., $x(k)$ transitions to $x(k+1)$. This is because the probability distribution $P$ for the next token is determined by the \textit{entire history of the text} $x(k)$, which includes both the initial text and all previously added tokens.}
\ColorA{Note that if we denote the number of tokens in $x_0$ as $k_0$, the total number of tokens in $x(k)$ is $k_0+k$.}

\ColorA{
The token predictor $G$ is a combination of a so-called an LLM and the softmax processing. Examples of LLMs include GPT-2 \cite{Brown20_GPT2} and Llama~3 \cite{Dubey24}.
The token distribution vector $P$ is derived by applying the softmax function to the LLM's output, as follows:
\begin{align}\begin{split}
    G: P[t] &= \mathrm{softmax}(\texttt{LLM}(x)[t]/T) \\
    &= \frac{
    \exp(\texttt{LLM}(x)[t]/T)
    }{
    \sum_{t'=1}^N \exp(\texttt{LLM}(x)[t']/T)
    },\quad t\in\TokenSet,
\label{E:P.TokenPredictorDetail}
\end{split}\end{align}
where $\texttt{LLM}:\TextSet\to\RR^N$ is the LLM, and $T>0$ is a hyperparameter called temperature. 
The token distribution vector $P$ has elements within the range $(0,1)$ and its elements sum up to $1$, while the pure output of the LLM does not. 
}

There are various methods for token selection, including greedy search and multinomial sampling.
Greedy search selects a token with the highest probability in $P$, i.e., $t^*=\arg\max_t P[t]$,
and the text-generation system \eqref{E:P.TextGenerationSystem} with greedy search is a \textit{deterministic} system.
Multinominal sampling randomly selects a token based on the given token distribution $P$, i.e., $\PP_{t^*\sim C(P)}[t^*=t]=P[t]$,
and the text-generation system \eqref{E:P.TextGenerationSystem} with greedy search is reduced to a \textit{stochastic} system.
In this paper, we employ multinomial sampling as the token selector.

\begin{figure}
    \centering
    \includegraphics[width=1\linewidth]{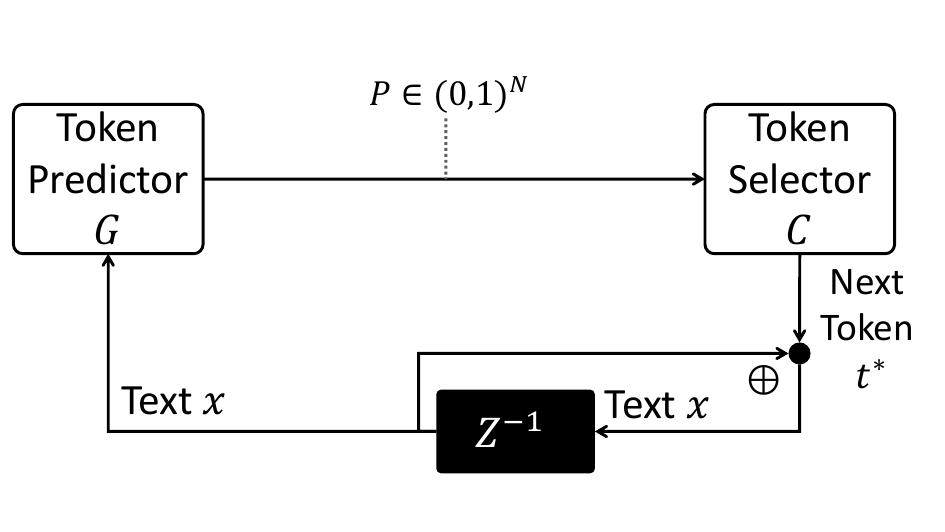}
    \caption{Text-generation system}
    \label{F:M.NominalTextGenerationSystem}
\end{figure}

The overall structure of the text-generation system \eqref{E:P.TextGenerationSystem} is shown in Fig.~\ref{F:M.NominalTextGenerationSystem}. \ColorA{In this figure, the blocks of $G$, $C$, and $Z^{-1}$ represent the token predictor, token selector, and time delay, respectively.}%

\section{LLM wigh CBF-Based Safety Filter}\label{Main:M}
This section presents the control-based alignment of text-generation systems and their detailed implementation.

The LLM alignment discussed in this paper aims to ensure desirable text generation by weak intervention to the output of the token predictor $G$ in text-generation system \eqref{E:P.TextGenerationSystem}.
To clarify the meaning of ``desirable'', we let the desirable and undesirable text sets be $\SafeSet\subseteq\TextSet$ and \ColorA{$\DangerSet = \TextSet \setminus \SafeSet \subseteq\TextSet$}, respectively, based on the respective alignment goals.
\begin{example}
    Consider that the alignment goal is set to generate texts with positive contexts.
    Then, $\SafeSet$ is the set of positive-context texts, and $\DangerSet$ is the set of non-positive-context texts.
    The details are seen in Section~\ref{Main:E}.
\end{example}

\begin{figure}
    \centering
    \includegraphics[width=1\linewidth]{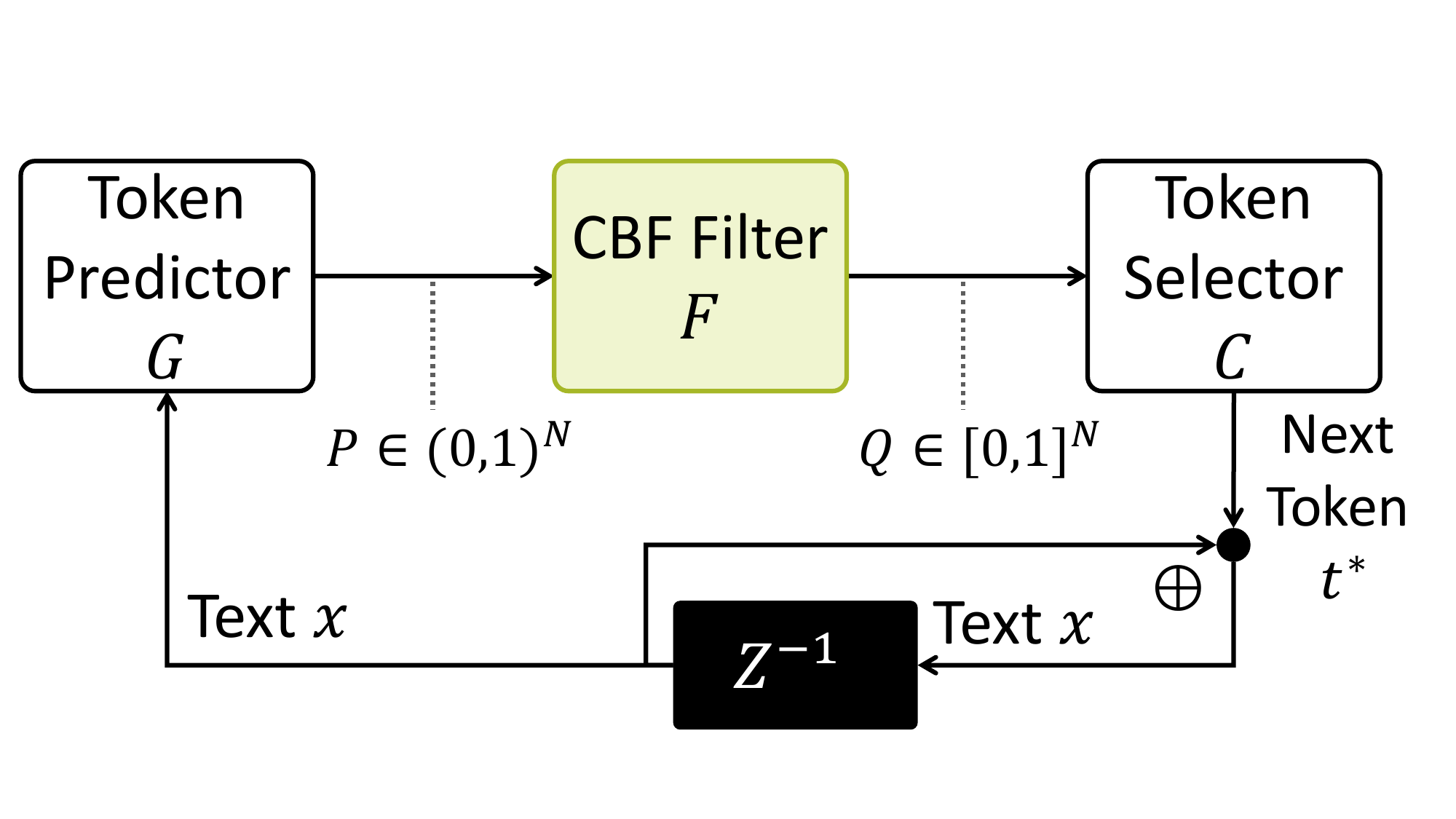}
    \caption{Proposed text-generation system with CBF filter (CBF-LLM)}
    \label{F:M.CBFLLMStructure}
\end{figure}

The presented text-generation system, including an LLM and a safety filter for alignment, is constructed based on the CBF described in Subsection~\ref{Main:P.CBF}.
The overall system is called CBF-LLM and its structure is shown in Fig.~\ref{F:M.CBFLLMStructure}.

CBF-LLM extends the nominal text-generation system shown in \eqref{E:P.TextGenerationSystem} by adding the CBF filter (green box) between the token predictor $G$ and the token selector $C$.
The CBF filter manipulates the token distribution $P$ to satisfy the user-defined alignment goal.
In the same manner as \eqref{E:P.TextGenerationSystem}, the blocks of $G$, $C$, and $Z^{-1}$ represent the token predictor, token selector, and time delay, respectively.


\subsection{Language-Constraint Function}

The defining feature of the CBF-LLM is the presence of CBF filter $F$, which filters $P$ to generate the modified token distribution $Q\in[0,1]^N$.
The CBF filter $F$ is designed by using the function $h:\TextSet\to\RR$ satisfying
\begin{align}\begin{cases}
    h(x)\ge0 , & x\in\SafeSet, \\
    h(x)<0 ,& x\in\DangerSet.
\label{E:M.LCF}
\end{cases}\end{align}
The function $h$ is called the ``language-constraint function'' (L-CF), whose role is the same as the constraint function presented in Subsection~\ref{Main:P.CBF}.
Note that the L-CF $h$ needs to distinguish between the desired and undesired texts accurately.
Furthermore, the value of L-CF is expected to indicate how desirable or undesirable the text is.

Generally, building the L-CF that perfectly distinguishes between desirable and undesirable texts is challenging.
One can construct the L-CF using existing text classification models.
An example of L-CF is provided as follows.
\begin{example}\label{Example:M.LCF}
    We apply a sentiment analysis RoBERTa model \footnote{\texttt{cardiffnlp/twitter\_roberta\_base\_sentiment\_latest} \cite{Loureiro22}} as the internal model of the L-CF.
    The RoBERTa model is originally trained to classify the context of given texts into three labels: negative, natural, or positive.
    Let \ColorA{$[s_-(x(k)) ~ s_\pm(x(k)) ~ s_+(x(k))]^\top \in [0,1]^3$} denote the softmax output of the RoBERTa model with respect to a given text $x$.
    It follows that \ColorA{$s_-(x(k))$, $s_\pm(x(k))$, and $s_+(x(k))$} represent the score of negative, neutral, and positive, respectively.
    Then, the L-CF $h$ is constructed as follows:
    \ColorA{
    \begin{align}
        h(x(k)) = s_+(x(k)) - \max(s_-(x(k)), s_\pm(x(k))) .
    \end{align}%
    }%
    This L-CF $h$ outputs a positive value when the positive score is greater than both negative and neutral scores,
    while it outputs a negative value when either the negative or neutral score is greater than the positive score.
    In other words, the sets $\SafeSet$ and $\DangerSet$, which correspond to the L-CF constructed above, render the texts with positive and non-positive contexts, respectively.
    Note that the RoBERTa model used in this L-CF is originally trained for evaluating whole texts, while it is used with mid-texts in this example.
    This may deteriorate the accuracy of the evaluation.
    \ColorA{For example, the model accurately evaluates the text input ``It is a nice day.'', but may not accurately evaluate the text input ``It is a''.}%
\end{example}

\subsection{CBF Filter and CBF-LLM}
The CBF filter $F:(0,1)^N\to[0,1]^N$ allows only tokens that meet its conditions to pass through and does not allow tokens that do not. The detailed realization of the CBF filter $F$ is given as follows:
\ColorB{
\begin{align}\begin{split}
    & F:Q(k)[t] \propto \\
    & \begin{cases}
        P(k)[t] ,& h(x(k)\oplus t) \ge \gamma h(x(k))\\
        0 ,& \text{else},
    \end{cases}
    \quad t\in\TokenSet,
    \label{E:M.CBFFilter}
\end{split}\end{align}
}%
where \ColorB{$\gamma\in[0,1]$} is a hyperparameter.
This formulation is a modified form of the discrete-time CBF inequality, as shown in \eqref{E:P.DiscreteTimeCBFConstraint}.
In \eqref{E:M.CBFFilter}, the probability of the token is set to $0$ unless the token satisfies the following CBF inequality:
\ColorA{
\begin{align}
    h(x(k)\oplus t) \ge \gamma h(x(k)),
    \label{E:M.CBFConstraint}
\end{align}
}%
which guarantees that the generated text $x$ always satisfies that $x\in\SafeSet$.
\begin{remark}
    The CBF inequality (\ref{E:M.CBFConstraint}) not only assesses whether $x\oplus t,t\in\TokenSet$ is undesirable or desirable but also takes into account how much the value of $h(x\oplus t)$ moves in the negative direction compared to the value of $h(x)$.
    Even if $x\oplus t$ is desirable, i.e., $h(x\oplus t)\ge0$, if the value of $h(x\oplus t)$ decreases significantly compared to $h(x)$, the probability of the token $t$ is set to 0.
    This conservative behavior aims to exclude not only tokens that immediately become undesirable but also tokens that ``give the conversation a dubious tone''.
\end{remark}

The output of the CBF filter, $Q$, needs to be normalized, and its elements sum up to $1$.
In addition, the CBF filter aims at user-desired text generation by weak interventions to the original token distribution $P$. 
To this end, we formulate the following optimization problem:
\ColorB{
\begin{subequations}
\begin{empheq}[left={
    \empheqlbrace
}]{alignat=1}
    \min_Q &~ \DKL[Q || P(k)] ,\label{E:M.CBFFilterProb.Obj}  \\
    \text{s.t.}
    &~ Q \ge \mathbf{0}, \label{E:M.CBFFilterProb.PositiveConst} \\
    &~ \mathbf{1}^\top Q = 1, \label{E:M.CBFFilterProb.SumConst} \\
    &~ \PP_{t\sim C(Q)}[h(x(k)\oplus t) \ge \gamma h(x(k))] = 1, \label{E:M.CBFFilterProb.CBFConst}
\end{empheq}
\label{E:M.CBFFilterProb}
\end{subequations}
}%
where $\DKL[Q||P]$ is Kullback-Leibler divergence between $Q$ and $P$.
The constraints (\ref{E:M.CBFFilterProb.PositiveConst}) and (\ref{E:M.CBFFilterProb.SumConst}) are imposed because $Q$ needs to be a distribution vector.
The constraint (\ref{E:M.CBFFilterProb.CBFConst}) requires that the $t$ selected based on $Q$ \textit{always} satisfy the CBF constraint (\ref{E:M.CBFConstraint}). \ColorA{The constraint is mathematically severe, and can be relaxed by replacing the probability of one with $\delta$, where $\delta\in[0,1)$ is an acceptable failure tolerance.}%

The CBF filter (\ref{E:M.CBFFilter}) is provided as follows:
\begin{theorem}\label{Theorem:M.CBFFilterProbMinimizer}
    The minimizer \ColorA{$Q(k)$} of the optimization problem (\ref{E:M.CBFFilterProb}) is provided by following:

    \begin{subequations}
    \begin{empheq}[left={\empheqlbrace}]{alignat=1}
    & P'(k)[t] =
    \begin{cases}
        P(k)[t] ,& \text{\small{The constraint \eqref{E:M.CBFConstraint} holds}}  ,\\
        0 ,& \text{else},
    \end{cases}
    \label{E:M.CBFFilterFirstStep}\\
    & Q(k)[t] = \frac{P'(k)[t]}{\sum_{i=1}^N P'(k)[i]},
    \label{E:M.CBFFilterSecondStep}\\
    & t\in\TokenSet. \notag
    \end{empheq}
    \end{subequations}
    
\end{theorem}
\noindent The proof of the \ColorA{theorem} is given in Appendix~\ref{Appendix:M.Proof}.

Top-K sampling is applied in the CBF filter $F$ to improve the computational efficiency. The top-K sampling only processes fewer elements than $N$ elements of the target $P$.
The algorithm of the CBF filter with top-K sampling is shown in Algorithm~\ref{Algorithm:TopK_SSA_CBFLLM}.

\begin{algorithm}[h]
\caption{CBF-LLM with top-K sampling}
\label{Algorithm:TopK_SSA_CBFLLM}
\begin{algorithmic}[1]
    \REQUIRE $P \in(0,1)^N$ : token distribution from the token predictor $G$.
    \REQUIRE $x \in\TextSet$ : current text.
    \REQUIRE $h:\TextSet\to\RR$ : the constraint function \eqref{E:M.LCF}.
    \REQUIRE \ColorB{$\gamma\in[0,1]$} : CBF's hyperparameter.
    \REQUIRE $K$ : the top-k parameter.

    \COMMENT{Initialization}
    \STATE $\AllowedSet \gets \phi$ : allowed set, a set of tokens that satisfy the CBF inequality.
    \STATE $P' \in[0,1]^N \gets 0_N$
    \STATE $T \in\{1,\ldots,N\}^N \gets \texttt{argsort}(P)$ : sort the indexes of $P$ in descending order.

    \COMMENT{Collect $K$ allowed tokens.}
    \STATE $i \gets 1$
    \WHILE{$|\AllowedSet|<K$}
        \STATE $t \gets T[i]$ : the token with $i$-th highest probability.
        \IF{\ColorB{$h(x\oplus t) \ge \gamma h(x)$ }: CBF inequality \eqref{E:M.CBFConstraint} holds,}
            \STATE $\AllowedSet \gets \AllowedSet \cup \{t\}$ : append the token to the allowed set.
            \STATE $P'[t] \gets P[t]$
        \ENDIF
        \STATE $i \gets i + 1$
    \ENDWHILE

    \COMMENT{Make a modified token distribution.}
    \STATE $Q=P' / \sum_{t\in\TokenSet} P'[t]$
    \STATE Select the next token $t^*$ based on $Q$.
\end{algorithmic}
\end{algorithm}

\noindent Note that the text generation is done by iteratively performing the Algorithm~\ref{Algorithm:TopK_SSA_CBFLLM}

\subsection{Extension to Multi-Step Ahead Method}
A drawback of the CBF-LLM system presented in Fig.~\ref{F:M.CBFLLMStructure} is that the intervention strategy may be too conservative: it may filter out texts that appear undesirable initially but become desirable when read to the last, e.g., ``You are clumsy, but you have high aspirations!''.

To overcome this drawback of the CBF-LLM system with \textit{single-step ahead} token prediction, we extend the CBF-LLM system with \textit{multi-step ahead} token prediction.
Let $H\in\{1,2,\ldots\}$ denote the prediction horizon, and further let $y$ denote the sequence composed of $H$ tokens, i.e., $y=[$``you''$, $``are''$,\ldots,t_H$].
At each time, the multi-step ahead method collects $K\in\{1,2,\ldots\}$ candidates of $H$-token sequences $y_1,y_2,\ldots,y_K$ generated from the baseline LLM that continues from $x(k)$
such that the CBF inequality \eqref{E:M.CBFConstraint} holds, i.e., $h(x\oplus y)\ge \gamma h(x)$.
The next $H$-token sequence is selected from these candidates according to the distribution $\PP[y \mid x(k)]$ derived by the baseline LLM.
The probability of the presence of an $H$-token sequence $y=[t_1,\ldots,t_H]$ from $x$ is as follows:
\begin{align}
    \PP[y\mid x] = \Pi_{h=0}^{H-1} G(x\oplus t_1 \oplus \cdots \oplus t_h)[t_{h+1}].
\end{align}
The detailed pseudo-code is presented in Algorithm~\ref{Algorithm:MSA_CBFLLM}.

\begin{algorithm}[h]
\caption{CBF-LLM with multi-step ahead}
\label{Algorithm:MSA_CBFLLM}
\begin{algorithmic}[1]
    \REQUIRE $G :\TextSet\to(0,1)^N$ : the token predictor.
    \REQUIRE $x \in\TextSet$ : current text.
    \REQUIRE $h:\TextSet\to\RR$ : the constraint function \eqref{E:M.LCF}.
    \REQUIRE \ColorB{$\gamma\in[0,1]$} : CBF's hyperparameter.
    \REQUIRE $H\in\{1,2,\ldots\}$ : the prediction horizon.
    \REQUIRE $K$ : sample size.
    
    \COMMENT{Initialization}
    \STATE $\mathcal Y \gets [~]$ : list of candidate $H$-token sequences. 
    \STATE $Q \gets [~]$ : list of probability of each candidate $H$-token sequence.
    \STATE $k \gets 0$ : counter of candidate $H$-token sequences.
    
    \COMMENT{Collect $H$-token sequences as candidates.}
    \WHILE{$k<K$}
        \STATE $y \gets$``'' : empty text.
        \STATE $q \gets 1$
        \FOR{$H$ times}
            \STATE $t^* \gets$ select the next token based on $G(x\oplus y)$.
            \STATE $q \gets q \ G(x\oplus y)[t^*]$
            \STATE $y \gets y \oplus t$ : extend the candidate sequence.
        \ENDFOR
        \IF{\ColorB{$h(x\oplus y)\ge \gamma h(x)$} : CBF inequality \eqref{E:M.CBFConstraint} holds,}
            \STATE $\mathcal Y$\texttt{.append}($y$) : now $y$ has $H$ tokens.
            \STATE $Q$\texttt{.append}($q$) : now $q$ is the probability that $y$ would follow $x$, i.e., $q=\PP[y\mid x]$.
            \STATE $k \gets k+1$ : increment the counter.
        \ENDIF
    \ENDWHILE

    \COMMENT{Extend the text.}
    \STATE Select the next $H$-token sequence from its candidates $\mathcal Y$ based on $Q$.
\end{algorithmic}
\end{algorithm}

\subsection{Previous Works}
There are various works addressing alignment by intervening in the output of the LLM, either from a representational perspective \cite{Cao21, Keskar19} or a semantic one \cite{Joshua24_ParserDriven, Mudgal24_ControlledDecoding, Xu24_SafeDecoding, Kevin21_FUDGE, Huang24_DeAL, Li24_InferenceTimeIntervention}.
In particular, the work \cite{Mudgal24_ControlledDecoding} proposes an intervention-based alignment method with multi-step ahead prediction, named Blockwise best-of-$K$ method.
The method selects the best of $K$ candidates of $H$-token sequences generated from the baseline LLM without imposing any constraints.
The common feature of these previous works is to intervene in the token probability based on a reward of generated text.
On the other hand, CBF-LLM, our proposed system, intervenes in the token probability based on the \textit{change} in reward caused by adding a new token, rather than just the generated text.

\section{Experiments}\label{Main:E}
In this section, we implement the CBF-LLM with Llama~3 and a RoBERTa model and verify the CBF-LLM's alignment ability, the number of interventions, generation time, and output quality.
In the experiments, we commonly employ Llama~3~8b \cite{Dubey24}, a pre-trained LLM, as the model for the token predictor $G$.

\subsection{Positive Text Generation}\label{Main:E.PTG}
The alignment goal is to ensure that the CBF-LLM system, illustrated in Fig.~\ref{F:M.CBFLLMStructure}, produces texts with ``positive'' contexts.
To this end, we let $\SafeSet$ and $\DangerSet$ denote the set of positive texts and non-positive texts, respectively.
We employ a RoBERTa model\footnote{\texttt{cardiffnlp/twitter\_roberta\_base\_sentiment\_latest}} to construct the L-CF $h$.
This RoBERTa model was originally trained to classify sentences into three labels: negative, neutral, or positive.
The L-CF outputs a positive value when the sentiment of the text $x$ is positive.
The detail is presented in Example~\ref{Example:M.LCF} in Section~\ref{Main:M}.
The resulting text-generation system would be controlled to generate positive content.

We use the Reddit dataset, \texttt{reddit-corpus-small} \cite{Convokit18} to collect the initial texts to be input for the text-generation system.
From the Reddit dataset, we randomly chose 50 utterance texts that satisfy the following three conditions:
\begin{enumerate}
    \item The text has more than 10 tokens.
    \item The text in which the L-CF $h$ indicates positive for the first 5-token text.
    \item The text in which the L-CF $h$ indicates negative for the generated text by the original Llama~3 model without any control gives the first 5-token text. In other words, the text of the first 5-token potentially results in a non-positive generation.
\end{enumerate}
We extract only the first 5 tokens from the selected utterance texts and use them as the initial texts $x_0$.
We set the temperature as $T=1$, the top-K value as $K=30$, and the maximum number of new tokens as 30.

In the experiments with text generation by CBF-LLM, as shown in Fig.~\ref{F:M.CBFLLMStructure}, we varied the hyperparameter \ColorB{$\gamma$} from $0.0$ to $1.0$.
We also implemented ``No Intervention'' case as a comparative baseline.
The No Intervention case performs the nominal text generation shown in \eqref{E:P.TextGenerationSystem}, without any intervention filters.

\begin{figure}
    \centering
    \includegraphics[width=1\linewidth]{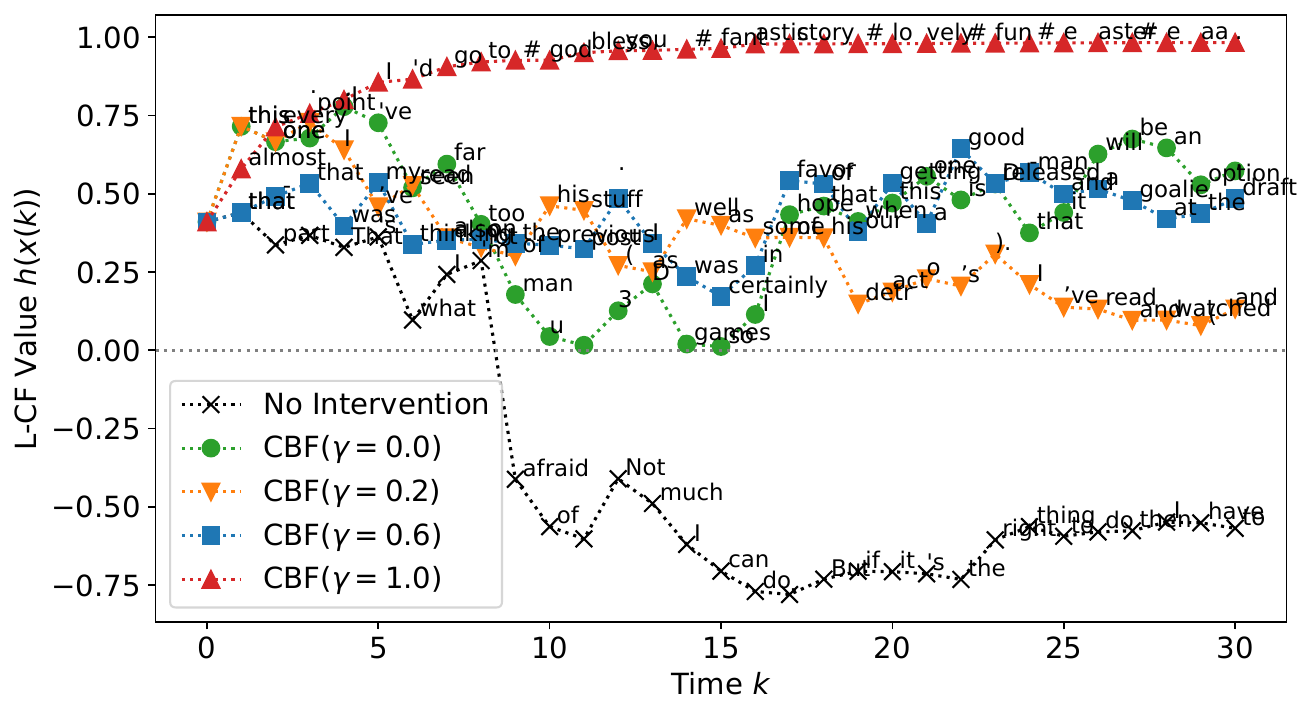}
    \caption{\ColorA{L-CF trajectory of each filter}}
    \label{F:E.PTG.LCFTrajectories}
\end{figure}

The trajectory of L-CF $h(x(k)), k\in\{1,2,\ldots\}$ for a generated text sample is shown in Fig.~\ref{F:E.PTG.LCFTrajectories}.
In the No Intervention case (black line), the generated text does not keep the positive L-CF value, implying the extent to which the generated text is undesirable.
On the other hand, in CBF filters, the L-CF values are kept positive during the generation, implying that the text generation system generates desirable content. \ColorA{Specifically, for the case $\gamma=1.0$ (red line), the L-CF $h(x(k))$ monotonically increased as $k$ progressed. This is because the generation was performed under the strict constraint that the CBF condition is written as $h(x(k+1)) \ge h(x(k))$, requiring the next text to be even more ``positive'' than the current one.}

In Fig.~\ref{F:E.PTG.LCFTrajectories}, the initial text $x_0$ was ``I agree with you on''. 
The generated texts are as follows: \footnote{\ColorA{The generated texts have a typo, showing ``manu’’ where it should be ``many’’. However, this is the direct output from the text-generation system and has been listed without modification.}}


\begin{framed}
\begin{itemize}
    \item No Intervention: \underline{I agree with you on} that part. \textcolor{red}{That's what I'm afraid of. Not much I can do.} But if it's the right thing to do then I have to
    \item CBF(\ColorB{$\gamma=0.0$}): \underline{I agree with you on} this one. I've seen far too manu 3D games so I hope that when this one is released that it will be an option
    \item CBF(\ColorB{$\gamma=0.2$}): \underline{I agree with you on} this one. I’ve read a lot of his stuff (as well as some of his detracto’s). I’ve read and watched (and
    \item CBF(\ColorB{$\gamma=0.4$}): \underline{I agree with you on} some of the above, especially \#4. I’ve been thinking a lot about the church lately. I’ve been attending St. Joseph’s parish.
    \ColorA{\item CBF($\gamma=1.0$): \underline{I agree with you on} almost every point. I'd go to \#god bless you, \#fantastic story. \#lovely \#fun \#easter \#eaa.}
\end{itemize}
\end{framed}
In the No Intervention case, we observe non-positive contexts in the generated texts (red text), whereas in the CBF cases, the method eliminates non-positive contexts.
\ColorA{However, when $\gamma$ is high, for example $\gamma=1.0$, the output text may contain expressions that appear unnatural in natural language, such as ``\#eaa.''. }

\begin{figure}
    \centering
    \includegraphics[width=1\linewidth]{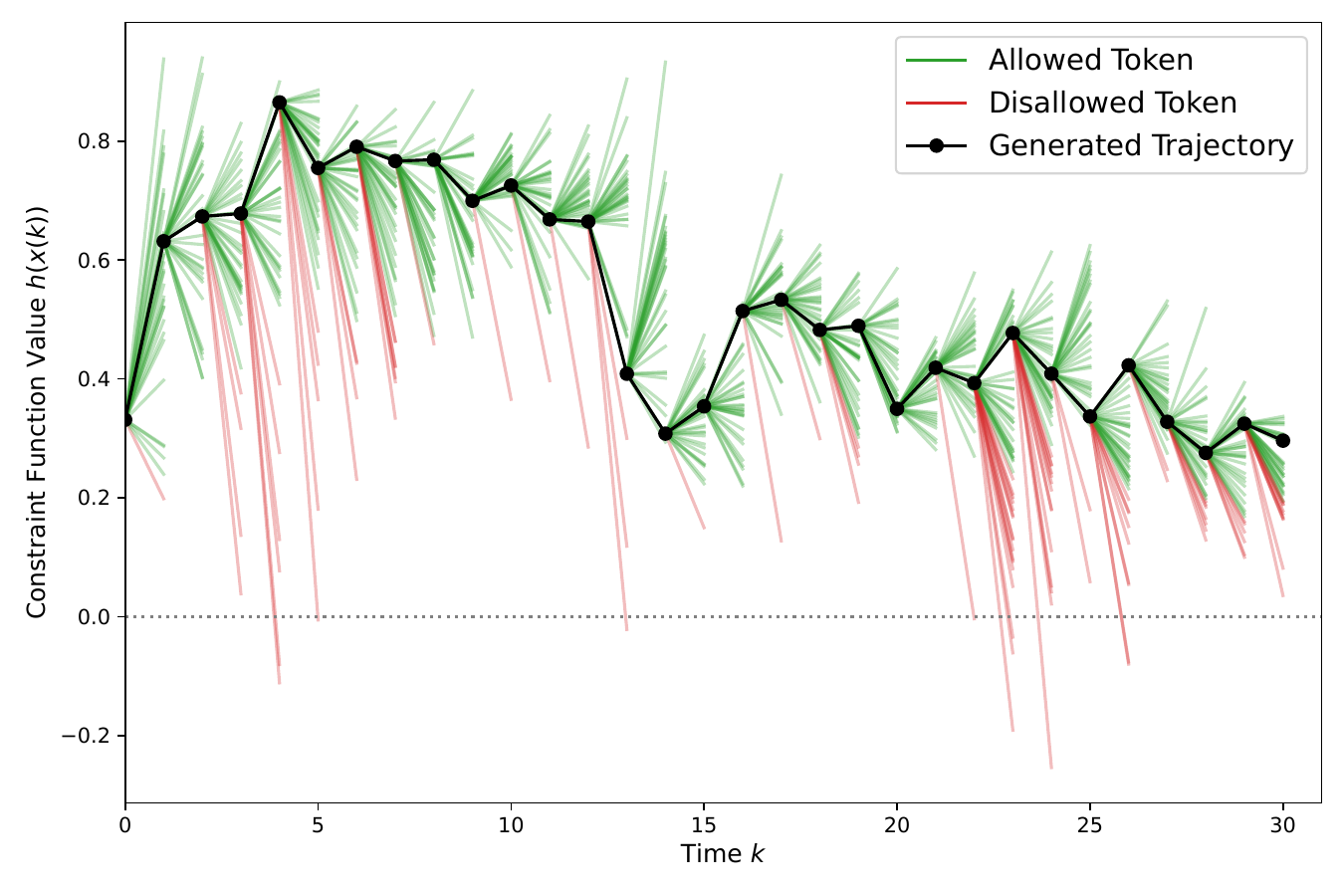}
    \caption{Predicted L-CF trajectory}
    \label{F:E.PTG.PredLCFTrajectory}
\end{figure}

The predicted possible trajectories of the CBF filter is shown in Fig.~\ref{F:E.PTG.PredLCFTrajectory}.
In CBF-LLM, the CBF filter sorts tokens into those that satisfy the CBF inequality \eqref{E:M.CBFConstraint} and those that do not.
The figure shows that the CBF filter prevents L-CF values from becoming negative of decreasing more rapidly than the current value.
Note that we do not show the trajectories for all tokens, but only for tokens investigated by top-K sampling are displayed.

We evaluated the generated texts from various aspects.
Recall that this paper aims to ensure desirable text generation by weakly intervening in the output of LLMs.
To evaluate the intervention weakness, we use the number of disallowed tokens per generation and naturalness. The average naturalness is evaluated by G-Eval \cite{Liu23_GEval}. 
Additionally, the positiveness of the generated texts, which was the control objective of this experiment, was assessed using G-Eval.
G-Eval is a method for evaluating the quality and task compliance of generated texts \ColorA{by using other LLMs}. The prompts used for evaluating naturalness and positiveness are shown in Appendix~\ref{Appendix:E.GEval}.
\ColorA{Furthermore, to compare the performance of text generation, we measure the average time the text-generation system takes to generate one token.}

The results are shown in Table~\ref{T:E.PTG.EvaluationResults}.
In the intervention cases, the naturalness score was the highest at \ColorB{$\gamma=0.4$}, the positiveness score was the highest at \ColorB{$\gamma=1.0$}, and the number of the disallowed tokens was the lowest at $\gamma=0.2$.
These results reveal a trade-off between task quality, displayed by positiveness, and intervention weakness, displayed by naturalness and disallowed tokens, and the trade-off is calibrated by the hyperparameter \ColorB{$\gamma$}.
\ColorA{The ``Time per Token'' column in the table shows the average time the system takes to generate one token. The text-generation system with CBF is about $0.03$ seconds longer than the baseline system. This is due to the evaluation of the L-CF $h$ taking time.}
\ColorB{The number of disallowed tokens at $\gamma=1.0$ becomes significantly larger than other $\gamma$. This is because the CBF constraint at $\gamma=1.0$ was strict, and few word sequences satisfy the constraint.}
Based on the results on naturalness, positiveness, the number of disallowed tokens, and the generation time, we see that the hyperparameter \ColorB{$\gamma$} should be chosen from within $(0,1)$ rather than choosing $1$, which is equivalent to the Blocklist.

\begin{table*}[t]
\centering
\caption{Evaluation Results}
\begin{tabular}{l|c|ccc}
\hline
    & \begin{tabular}{c}Number of  disallowed tokens\\per generation\end{tabular}
    & Naturalness 
    & Positiveness
    & \ColorA{Time per Token$~\mathrm{[s]}$} \\
\hline
    \ColorA{CBF($\gamma=1.0$)} & \ColorA{$1.05\times10^4$} & \ColorB{0.324} & \ColorB{\Strong{0.660}} & \ColorA{0.548}\\
    CBF($\gamma=0.8$) & 575 & \ColorA{0.327} & \ColorA{0.385} & \ColorA{0.140} \\
    CBF($\gamma=0.6$) & 335 & \ColorA{0.347} & \ColorA{0.350} & \ColorA{\Strong{0.135}} \\
    CBF($\gamma=0.4$) & 469 & \ColorA{\Strong{0.364}} & \ColorA{0.363} & \ColorA{0.137} \\
    CBF($\gamma=0.2$) & \Strong{299} & \ColorA{0.333} & \ColorA{0.356} & \ColorA{\Strong{0.135}} \\
\hline
    CBF($\gamma=0.0$) & 368 & \ColorA{0.358} & \ColorA{0.368} & \ColorA{0.141} \\
\hline
    No Intervention & 0 & \ColorA{0.375} & \ColorA{0.242} & \ColorA{0.114} \\
\hline
\end{tabular}



\label{T:E.PTG.EvaluationResults}
\end{table*}

\subsection{Extension to Multi-Step Ahead CBF-LLM}\label{Main:E.MSA}
We conduct an additional experiment to demonstrate the text generation by CBF-LLM with the multi-step ahead method.
In this experiment, we compare the multi-step ahead CBF-LLM and the Blockwise best-of-$K$ \cite{Mudgal24_ControlledDecoding}.
The alignment goal, the baseline LLM, the L-CF, and the initial texts used in this experiment are the same as those in Subsection~\ref{Main:E.PTG}.

\ColorA{We evaluated the naturalness by G-Eval, the rate of generated texts that were not positive, and the generation time}.
The results are shown in Table~\ref{T:E.MSA.EvaluationResults}.
At each element, the left\ColorA{, middle, and right side values display the non-positive generation rate, naturalness, and the average time the text-generation system takes to generate one token, respectively.}
We can see that the naturalness is higher in the multi-step ahead CBF-LLM compared to the Blockwise best-of-$K$ method.
Notably, when the sample size $K$ is small, the Blockwise best-of-$K$ had a relatively high rate of non-positive text generation.
The Blockwise best-of-$K$ method does not disallow the user-undesired outputs, which may lead to user-undesired results.
In contrast, the multi-step ahead CBF-LLM did not produce any undesirable text for any value of $K$ due to the safety filter.

Given the practical need to reduce $K$ due to some reason, such as computational efficiency, the multi-step ahead CBF-LLM has potential in scenarios where avoiding undesirable text is guaranteed.

\ColorA{The multi-step ahead CBF-LLM also revealed challenges. Specifically, it takes longer generation time than the \textit{single-step} version of CBF-LLM, presented in Subsection~\ref{Main:E.PTG}. This is because the generation process requires producing a larger number of candidate token sequences.}

\begin{table*}[t]
    \centering
    \caption{Non-Positive Generation Rate, Naturalness, \ColorA{and Time per Token}}
    \begin{tabular}{c|cc}
\hline
    \begin{tabular}{c} Sample Size \\$K$ \end{tabular}
    &
    \begin{tabular}{c} Multi-Step Ahead CBF-LLM\\($H=3,\alpha=0.8$) \end{tabular}
    &
    \begin{tabular}{c} Blockwise best-of-$K$ \cite{Mudgal24_ControlledDecoding}\\($H=3$)\end{tabular}\\
    \hline 
         2 & 
            $\Strong{0.00}/\Strong{0.722}/ \ColorA{0.157~\mathrm{s}}$ &
            $0.30/0.592/ \ColorA{0.309~\mathrm{s}}$\\
         4 & 
            $\Strong{0.00}/\Strong{0.718}/\ColorA{0.333~\mathrm{s}}$ &
            $0.02/0.695/ \ColorA{0.655~\mathrm{s}}$\\
         5 & 
            $0.00/\Strong{0.727}/ \ColorA{0.419~\mathrm{s}}$ & 
            $0.00/0.701/ \ColorA{1.05~\mathrm{s}}$\\
    \hline
    \end{tabular}
    \label{T:E.MSA.EvaluationResults}
\end{table*}

\section{Concluding Remarks}\label{Main:C}
This paper proposed the control-based LLM alignment framework, called CBF-LLM.
This framework utilizes the control barrier function (CBF) to ensure the safety of physical objects, such as the collision avoidance function in assisted driving vehicles.
Based on an analogy between the control theory and the LLM alignment task,
we employed the CBF-based safety filter to ensure that the text-generation system generates desirable content.
The key feature of CBF-LLM is that the CBF filter can be attached to the baseline LLM in an add-on manner: 
it intervenes in the output of the baseline LLM without any additional training of LLMs.
This paper also presented the implementation of CBF-LLM by Llama~3 and a sentiment analysis RoBERTa model to ensure that the text-generation system generates positive content.
The text-generation experiment showed that CBF-LLM outperforms the baseline method in terms of naturalness, positiveness, generation time, and the reliability of controlled decoding.
\ColorA{For the evaluation of the generated text quality, our experiments employed a method relying on another LLM. Future work should prioritize using more objective evaluation methodologies that reduce ambiguity.}%

In CBF-LLM, the key challenge is to effectively incorporate human feedback and existing evaluation models to reflect human preferences into the L-CF.
The design of L-CF is similarly challenging to construct a high-quality reward model in RLHF approaches.
The value of CBF filters lies in their ability to facilitate easy modifications. 
To illustrate this, we show two scenarios:
In a scenario, consider that an aligned LLM is developed and integrated into a service system. 
Suppose that an ethical or other critical issue is discovered with the original data used for alignment.
Then, it becomes challenging to remove the influence of the data from the LLM using RLHF-based methods, such as unlearning \cite{Isonuma24_Unleaning}. 
This can lead to the suspension of the service system.
In contrast, in CBF-LLM, an add-on type alignment method, we can simply disable the CBF filter to maintain the service operation while modifying the specifications.
In the other scenario, consider an LLM initially trained or controlled to produce positive text.
Later, suppose that an additional requirement is added such as ensuring that the generated text is easy for children to comprehend. 
In CBF-LLM, we can independently design a readability CBF filter without modifying the existing positivity CBF filter, allowing the system to meet the updated requirements without having to retrain the entire LLM. 
This approach enables us to easily adapt to changing specifications and requirements.
In these scenarios, the CBF-LLM approach offers a significant advantage.

\bibliographystyle{ieeetr}
\bibliography{
References/Control,
References/LLM,
References/Experiments,
References/Other
}

\appendix

\subsection{Proof of \ColorA{Theorem}~\ref{Theorem:M.CBFFilterProbMinimizer}}
\label{Appendix:M.Proof}
This subsection proves the property of the CBF filter (\ref{E:M.CBFFilter}).
To this end, we define allowed and disallowed sets $\AllowedSet$, $\DisallowedSet$.
The allowed set $\AllowedSet$ is a set of tokens that hold the CBF inequality \eqref{E:M.CBFConstraint}, 
while the disallow set $\DisallowedSet$ is a set of tokens that do not.

Given $P$, we consider the optimization problem (\ref{E:M.CBFFilterProb}).
To satisfy the constraint (\ref{E:M.CBFFilterProb.CBFConst}), it is clear that the probability of every disallowed token $t,\DisallowedSet$ must be $0$.
Therefore, the constraint (\ref{E:M.CBFFilterProb.CBFConst}) is written as $Q[t]=0,\forall t\in\DisallowedSet$.
Now, the optimization problem (\ref{E:M.CBFFilterProb}) is rewritten as follows:
\begin{subequations}\label{E:M.CBFFilterProb1}
\begin{empheq}[left={
    \empheqlbrace
}]{alignat=1}
    \min_Q &~ \DKL[Q || P] ,
    \label{E:M.CBFFilterProb1.Obj}\\
    \text{s.t.}
    &~ Q[t] \ge 0, \quad\forall t\in\AllowedSet,
    \label{E:M.CBFFilterProb1.AllowedNonNegaConst}\\
    &~ Q[t] = 0, \quad\forall t\in\DisallowedSet,
    \label{E:M.CBFFilterProb1.DisallowedConst}\\
    &~ \sum_{t\in\AllowedSet} Q[t] = 1.
    \label{E:M.CBFFilterProb1.SumConst}
\end{empheq}
\end{subequations}
%

The KL divergence \eqref{E:M.CBFFilterProb1.Obj} is rewritten as $\DKL[Q||P] = \sum_{t\in\TokenSet} Q[t] \ln \frac{Q[t]}{P[t]}$.
Recall that $Q[t]=0,\forall t\in\DisallowedSet$, the KL divergence (\ref{E:M.CBFFilterProb1.Obj}) is reduced to
\begin{align}
    \sum_{t\in\AllowedSet} Q[t]\ln\frac{Q[t]}{P[t]} .
    \label{E:M.CBFFilterProb1.KLRewritten}
\end{align}


From here, we are focusing only on allowed tokens $t\in\AllowedSet$ on the KL divergence.

We next show that $Q[t]=0$ for some $t\in\AllowedSet$ cannot be the minimizer.
The gradient of KL divergence (\ref{E:M.CBFFilterProb1.KLRewritten}) is given by
\begin{align}
    \frac{\partial}{\partial Q[t]} \left[
    \sum_{t\in\AllowedSet} Q[t]\ln\frac{Q[t]}{P[t]}
    \right]
    =
    \ln\frac{Q[t]}{P[t]}+1
    ~,t\in\AllowedSet.
\end{align}
Note that the gradient goes $-\infty$ as $Q[t]\to+0$.
This implies that $Q[t]=0$ for some $t\in\AllowedSet$ is not the minimizer.
Therefore, the constraint (\ref{E:M.CBFFilterProb1.AllowedNonNegaConst}) is reduced to $Q[t] > 0,t\in\AllowedSet$.
Then, the optimization problem (\ref{E:M.CBFFilterProb1}) is further simplified as follows:
\begin{subequations}\label{E:M.CBFFilterProb2}
\begin{empheq}[left={
    \empheqlbrace
}]{alignat=1}
    \min_Q &~ \sum_{t\in\AllowedSet} Q[t]\ln\frac{Q[t]}{P[t]} , 
    \label{E:M.CBFFilterProb2.Obj}\\
    \text{s.t.}
    &~ Q[t] > 0, \quad\forall t\in\AllowedSet, 
    \label{E:M.CBFFilterProb2.AllowedPositiveConst}\\
    &~ \sum_{t\in\AllowedSet} Q[t] = 1.
    \label{E:M.CBFFilterProb2.SumConst}
\end{empheq}
\end{subequations}
%

The objective function \eqref{E:M.CBFFilterProb2.Obj} is convex to $Q[t],t\in\AllowedSet$ and has the minimizer, since the Hessian matrix is positive definite, i.e.,
\begin{align}
    H(Q[t],t\in\AllowedSet)
    = \mathrm{diag} \left\{
        \frac{1}{Q_\AllowedSet [1]} ,\ldots, \frac{1}{Q_\AllowedSet [|\AllowedSet|]}
    \right\}
    \succ 0,
\end{align}
where $Q_\AllowedSet[i]$ is the probability of $i$-th allowed token.

Now, we formulate the Lagrange function $L$ as follows:
\begin{align}
    L := 
        \sum_{t\in\AllowedSet} Q[t]\ln\frac{Q[t]}{P[t]}
        \ + \
        \lambda \left( \sum_{t\in\AllowedSet} Q[t]-1 \right),
\end{align}
where $\lambda$ is the Lagrange multiplier.
The minimizer of the optimization problem (\ref{E:M.CBFFilterProb2}) should satisfy the following equation:
\begin{align}
    \frac{\partial L}{\partial Q[t]} = 0 , \quad\forall t\in\AllowedSet.
\end{align}
For each allowed token $t\in\AllowedSet$, it follows that:
\begin{align}
    \frac{\partial L}{\partial Q[t]} 
    = \ln\frac{Q[t]}{P[t]} + 1 + \lambda = 0.
\end{align}
This implies that, $Q[t],t\in\AllowedSet$ is 
\begin{align}
    Q[t] = e^{-(1+\lambda)} P[t]  ,\quad\forall t\in\AllowedSet.
\end{align}
Since the coefficient $e^{-(1+\lambda)}$ is same for all allowed tokens $t\in\AllowedSet$, we see that \eqref{E:M.CBFFilterProb2.SumConst} holds for
$\lambda = \ln\sum_{t\in\AllowedSet}P[t]-1$ and this reduces the coefficient  $e^{-(1+\lambda)}$ to
\begin{align}
    Q[t] = \frac{P[t]}{\sum_{t\in\AllowedSet}P[t]}, \quad\forall t\in\AllowedSet.
\label{E:M.CBFFilterProof.0}
\end{align}

This concludes that the proposition holds,
meaning that the CBF filter, shown in \eqref{E:M.CBFFilter}, has the KL minimality under the CBF inquality (\ref{E:M.CBFConstraint}).
\hfill \qed

\subsection{Evaluation of Naturalness and Positiveness}\label{Appendix:E.GEval}
In the naturalness and positiveness evaluation by G-Eval framework \cite{Liu23_GEval}, we used \ColorA{GPT-5-mini} and the following prompt.
These scores are normalized by dividing the response values by 10.

The prompt for naturalness evaluation is as follows:

\textit{\small\\
Given the evaluation steps, return a JSON with two keys: 1) a `score` key ranging from 0 - 10, with 10 being that it follows the criteria outlined in the steps and 0 being that it does not, and 2) a `reason` key, a reason for the given score, but DO NOT QUOTE THE SCORE in your reason. Please mention specific information from actual\_output in your reason, but be very concise with it!
Evaluation Steps:
1. Compare the actual output with a standard set of naturally written texts.\\
2. Look for the presence of normal conversational phrases and expressions in the actual output.\\
3. Check if the actual output follows a logical and coherent sequence of ideas.\\
4. Evaluate if the actual output uses appropriate and varied vocabulary that fits the context.\\
\\
actual\_output : \texttt{Output text}\\
\\
**\\
IMPORTANT: Please make sure to only return in JSON format, with the "score" and "reason" key. No words or explanation is needed.\\
\\
Example JSON:\\
\{\{\\
    "score": 0,\\
    "reason": "The text does not follow the evaluation steps provided."\\
\}\}\\
**\\
\\
JSON:\\
"""\\
}

The prompt for positiveness evaluation is as follows:

\textit{\small\\
Given the evaluation steps, return a JSON with two keys: 1) a `score` key ranging from 0 - 10, with 10 being that it follows the criteria outlined in the steps and 0 being that it does not, and 2) a `reason` key, a reason for the given score, but DO NOT QUOTE THE SCORE in your reason. Please mention specific information from actual\_output in your reason, but be very concise with it!
Evaluation Steps:
1. Identify and note down all the positive words and phrases used in the given text.\\
2. Evaluate the frequency and distribution of these positive words/phrases throughout the text.\\
3. Assess the context in which these positive words/phrases are used, to ensure they are indeed contributing to a positive sentiment.\\
4. Compare the frequency, distribution, and context of positive words/phrases in the given text with those in other texts to determine its positivity level.\\
\\
actual\_output : \texttt{Output text}\\
\\
**\\
IMPORTANT: Please make sure to only return in JSON format, with the "score" and "reason" key. No words or explanation is needed.\\
\\
Example JSON:\\
\{\{\\
    "score": 0,\\
    "reason": "The text does not follow the evaluation steps provided."\\
\}\}\\
**\\
\\
JSON:\\
"""\\
}

\end{document}